\pdfoutput=1

\documentclass[11pt]{article}
\PassOptionsToPackage{table}{xcolor}
\usepackage{ACL2023}

\usepackage{times}
\usepackage{latexsym}

\usepackage[T1]{fontenc}

\usepackage[utf8]{inputenc}

\usepackage{microtype}

\usepackage{inconsolata}
\usepackage{enumerate} 
\usepackage{graphicx}
\usepackage{multirow}
\usepackage{enumitem}
\usepackage{paralist}
\usepackage{algorithm}
\usepackage{algpseudocode}
\usepackage{enumitem}
\usepackage{paralist}
\usepackage{amsmath, bm}
\usepackage{mismath}
\usepackage{tabularx}
\newcommand{\am}[1]{\textcolor{red}{#1 -- AM}}

\usepackage{marvosym}
\usepackage[table]{xcolor}
\usepackage{graphicx}
\usepackage{pgfplots}
\usepackage{pgfplotstable}
\usepackage{tikz}
\usepackage{mdframed} 
\definecolor{lightred}{rgb}{1, 0.7, 0.7}
\definecolor{lightblue}{rgb}{0.7, 0.7, 1}
\definecolor{darkred}{rgb}{0.6, 0, 0}
\definecolor{darkblue}{rgb}{0, 0, 0.6}

\pgfplotsset{compat=1.18}
\newmdenv[
  topline=false,
  bottomline=false,
  skipabove=\topsep,
  skipbelow=\topsep,
  leftline=true,
  rightline=true,
  linecolor=cyan,
  linewidth=2pt,
  innertopmargin=10pt,
  innerbottommargin=10pt,
  innerrightmargin=10pt,
  innerleftmargin=10pt,
  backgroundcolor=gray!10,
  roundcorner=10pt
]{stylishframe}

\title{Context Matters: Pushing the Boundaries of Open-Ended Answer Generation with Graph-Structured Knowledge Context}

\author{%
  Somnath Banerjee~$^\dagger$ 
  Amruit Sahoo~$^\dagger$ 
  Sayan Layek~$^\dagger$ 
  Avik Dutta~$^\dagger$\\
  \textbf{Rima Hazra}~$^\mp$
  \textbf{Animesh Mukherjee}~$^\dagger$\\
  $^\dagger$Indian Institute of Technology Kharagpur, India\\
  $^\mp$Singapore University of Technology and Design, Singapore\\
  \texttt{ \{som.iitkgpcse,sayanlayek2002\}@kgpian.iitkgp.ac.in}\\
 \texttt{ \{rima\_hazra\}@sutd.edu.sg} \\
 }

\begin{document}
\maketitle
\begin{abstract}
  \if{0}In the continuously advancing AI landscape, crafting context-rich and meaningful responses via Large Language Models (LLMs) is essential. Researchers are becoming more aware of the challenges that LLMs with fewer parameters encounter when trying to provide suitable answers to open-ended questions. To address these hurdles, the integration of cutting-edge strategies, augmentation of rich external domain knowledge to LLMs, offers significant improvements.\fi This paper introduces a novel framework that combines graph-driven context retrieval in conjunction to knowledge graphs based enhancement, honing the proficiency of LLMs, especially in domain specific community question answering platforms like AskUbuntu, Unix, and ServerFault. We conduct experiments on various LLMs with different parameter sizes to evaluate their ability to ground knowledge and determine factual accuracy in answers to open-ended questions. Our methodology \textsc{GraphContextGen} consistently outperforms dominant text-based retrieval systems, demonstrating its robustness and adaptability to a larger number of use cases. This advancement highlights the importance of pairing context rich data retrieval with LLMs, offering a renewed approach to knowledge sourcing and generation in AI systems. We also show that, due to rich contextual data retrieval, the crucial entities, along with the generated answer, remain factually coherent with the gold answer.
\end{abstract}

\section{Introduction}
In artificial intelligence, Large Language Models (LLMs)\cite{roberts-etal-2020-much, kaplan2020scaling} have revolutionized text understanding\cite{lian2023llmgrounded} and generation~\cite{wei2023chainofthought}. Despite their impressive capabilities, LLMs struggle in low-resource settings~\cite{chen2023exploring, guu2020realm}, are constrained by knowledge cutoffs, and often produce hallucinations~\cite{mckenna2023sources}. Additionally, managing the trade-off between quality and the vast number of parameters~\cite{xu2023compress} presents challenges, particularly for researchers with limited resources.\\
\begin{figure}[h]
\centering
\includegraphics[width=0.49\textwidth]{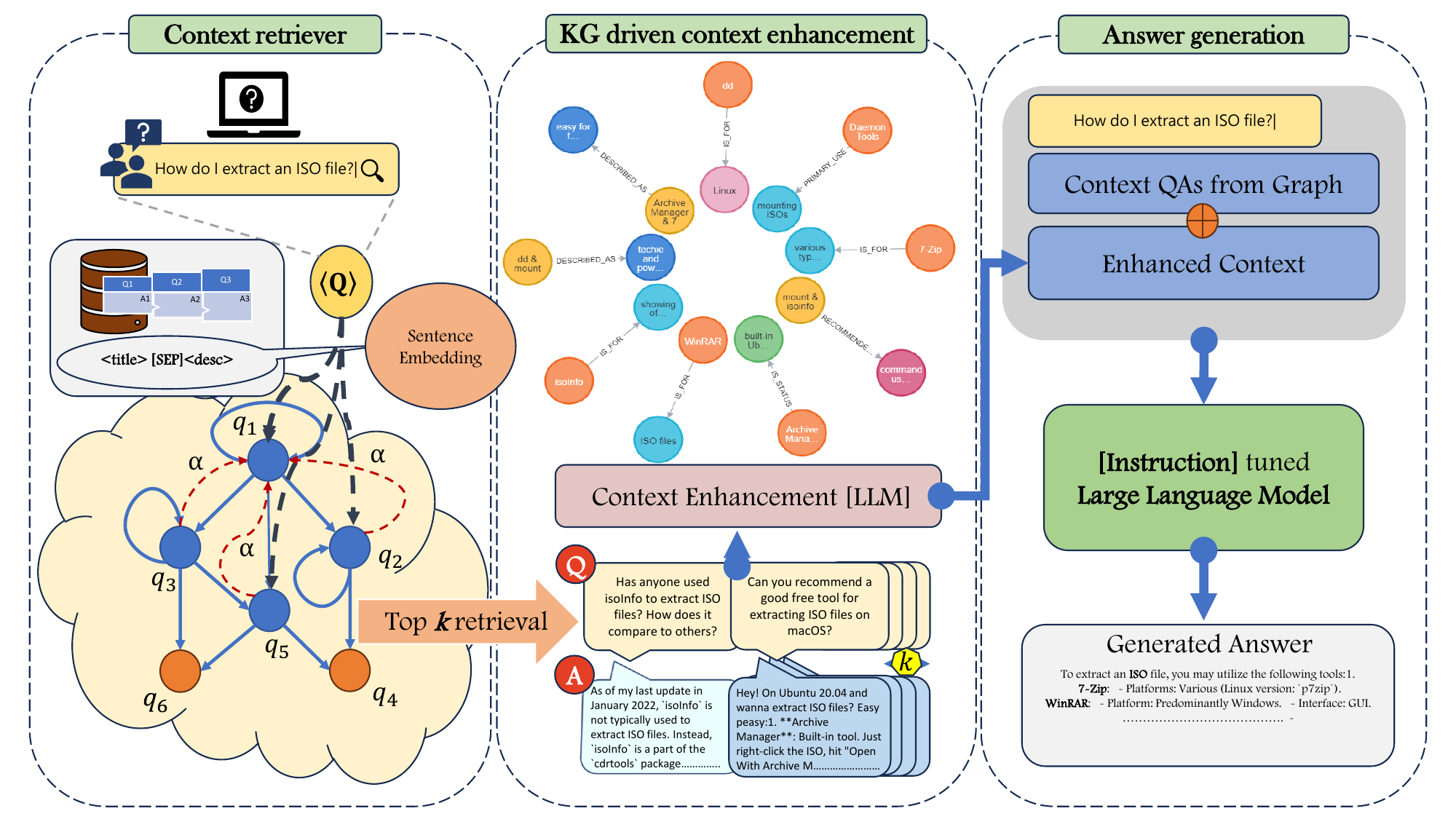}
\caption{\textsc{GraphContextGen} framework.} 
\label{fig2}
\vspace{-0.2cm}
\end{figure}
\noindent To overcome these limitations, new methods such as grounding LLMs~\footnote{https://techcommunity.microsoft.com/t5/fasttrack-for-azure/grounding-llms/ba-p/3843857} and Retrieval-Augmented Generation (RAG)~\cite{yu2023augmentationadapted} have been proposed. These approaches enable models to access external databases, enhancing their responses with current, detailed, and accurate information.\\
\noindent A critical aspect of effective knowledge grounding is the retrieval mechanism~\cite{lewis2021retrievalaugmented}. Traditional text-based retrieval methods are evolving to handle more complex questions, moving beyond simple keyword matching. Current techniques often struggle with determining optimal chunk sizes~\footnote{https://www.pinecone.io/learn/chunking-strategies/} for indexing and querying, leading to inconsistent results. Graph-based retrieval systems offer a solution by capturing intricate relationships through structured data, providing deeper semantic understanding and more contextually relevant results~\cite{zhang2021graphbased}. These systems adapt to evolving data, uncovering insights and forming connections among diverse entities.\\
\noindent This technique is vital in various applications, including dialogue systems~\cite{li2022knowledgegrounded}, open-domain question answering~\cite{lu2023structured}, and novelty-controlled paraphrasing~\cite{9978727}. For example, StackOverflow's OverflowAI~\footnote{https://stackoverflow.blog/2023/07/27/announcing-overflowai/} aims to refine search and enhance code and discussion platforms. Automated answer generation on community Q\&A platforms promises timely and accurate information, reducing errors and providing immediate knowledge access. Unlike past research, our study uniquely employs LLMs to generate tailored answers for these platforms.\\
\noindent We introduce the \textsc{GraphContextGen} framework, which combines graph-based retrieval with LLMs to enhance context and ensure factual accuracy. Our extensive experiments in low-resource domains like AskUbuntu\footnote{https://askubuntu.com/}, Unix~\footnote{https://community.unix.com/}, and ServerFault~\footnote{https://serverfault.com/} demonstrate the effectiveness and resilience of LLM-generated answers, even in specialized areas.

\noindent\textbf{Contribution}: The key contribution of this paper is as follows.
\begin{stylishframe}
\begin{compactitem}
\item We introduce \textsc{GraphContextGen}, a framework that integrates graph-based retrieval with context enhancement using a knowledge graph for CQA answer generation. This approach consistently outperforms previous SOTA methods. Additionally, instruction tuning with LLMs further improves performance (see Section~\ref{sec:method}).\\
\item We evaluate a range of recently released LLMs from 1.5B to 40B on CQA of low resource domains for answer generation in a zero shot setting (see Table~\ref{tab:zeroshot}).\\
\item In addition to automatic evaluation, we also perform evaluation based on human judgements and demonstrate that in both cases our proposed framework consistently outperforms all current SOTA text-based retrieval techniques (see Table~\ref{tab:maintable} and Section~\ref{sec:results}).\\
\item We conduct a detailed retrospective analysis to compare actual answers with those generated by our framework, focusing on their factual alignment. The generated answers typically align well with the actual ones (see Figure~\ref{fig:my_label}).
\end{compactitem}
\end{stylishframe}

\section{Related Work}
\if{0}
\noindent \textbf{Large language models}: Recent progress in LLMs~\cite{naveed:2023, douglas:2023} has positioned them as dominant successors to earlier transformer-based models for tackling generation tasks in natural language. Over the past several years, there has been a surge in introducing LLMs~\footnote{https://huggingface.co/spaces/HuggingFaceH4/\\open\_llm\_leaderboard} like GPT3~\cite{brown:2020}, Llama2~\cite{touvron:2023} etc. for this purpose. The newly introduced LLMs have demonstrated the ability to grasp and interpret context, subsequently producing human like responses. 
These models generate fluent and high quality answers very well for general questions~\cite{yang-etal:2023}. So, these models are often used as the base architecture for various generation based tasks such as long Q\&A~\cite{saadfalcon2023pdftriage, kamalloo-etal-2023-evaluating}, multiple choice Q\&A~\cite{robinson2023leveraging}, dialogue systems~\cite{deng2023prompting, hudeček2023llms, valvoda2022prompting, snell-etal-2022-context}, chatbot systems~\cite{lee2023prompted, 2023arXiv230105843W}. However, despite the high capability of these methods, they generate factually incorrect~\cite{shuster:2021} responses and are oftentimes not aligned with the actual question for domain specific knowledge. Generally, this issue of LLMs is called hallucination~\cite{dhuliawala2023, mündler2023}. Due to the cutoff knowledge, models sometime fail to provide faithful answer for a general question. To tackle this problem, recent studies~\cite{shuster:2021} proposed a method of augmenting retrieved knowledge from diverse sources as a context to LLMs.\\
\noindent\textbf{Retrieval augmented generation}: Retrieval-augmented generation (RAG)~\cite{manathunga:2023, ram:2023}
is a framework that merges the capabilities of large pretrained language models with the advantages of external retrieval or search mechanisms. Earlier, augmenting language models with retrieved knowledge had shown effective performance for knowledge-intensive tasks~\cite{Guu:2020}. 
It is particularly useful in scenarios where a language model needs to pull in specific factual information from an extensive corpus to generate relevant and accurate outputs. In recent studies, the researchers attempt to augment external knowledge through knowledge graph prompting~\cite{wang:2023}, few shot domain adaptation~\cite{krishna:2023}, retrieval-generation synergy~\cite{shao:2023}, knowledge graph-based subgraph retrieval augmented generation~\cite{kang:2023}, augmented adapter retriever (AAR)~\cite{yu2023augmentationadapted}.\\
\noindent\textbf{Instruction tuning}: Although LLMs have already shown impressive generation capabilities, oftentimes they cannot align the user's objective with its training objective~\cite{zhang:2023}. This is because while in general LLMs are trained to minimize the contextual word prediction error, users want to receive responses given some instructions of their choice. To mitigate that, various methods such as supervised fine tuning (instruction tuning)~\cite{vonwerra:2022}, RLHF~\cite{ziegler:2019, Stiennon:2020, ouyang:2022, gao:2022}, RLAIF~\cite{lee:2023} have been used to enhance the capabilities and controllability of LLMs. Also, parameter efficient techniques~\cite{peft} have been proposed to fine tune less number of model parameters and more adapter's parameters.

\noindent\textbf{Answer generation}: \fi

Over the years several approaches such as feature based methods~\cite{wang:2009, wangmanning:2010}, CNN~\cite{Severyn:2015, Rao:2017}, RNN~\cite{wang-nyberg-2015-long}, attention mechanism~\cite{tan-etal-2016,Santos2016AttentivePN} have been proposed for answer selection and summarization.
Some recent research focuses on summarizing diverse content on StackOverflow~\cite{chengran2022answer}, using AnswerBot--an answer summary generator~\cite{Xu:2017}, Opiner--which summarizes API reviews~\cite{Uddin:2017}, extracting key sentences to guide developers on StackOverflow~\cite{nadi2019essential}, and multi-document summarization~\cite{xu-lapata-2020-coarse}. In the realm of answer summarization, numerous studies~\cite{ganesan2010opinosis,Naghshzan_2021} harness graphical structures, leverage existing graph-based summarizers~\cite{mihalcea-tarau-2004-textrank, Erkan:2004, kazemi-etal-2020-biased}, and employ graph-centric measures. In the age of LLMs, some research has centered around controlled summary generation via effective keyword-based prompting~\cite{he-etal-2022-ctrlsum}.
\section{Dataset}
\vspace{-0.2cm}
\begin{table}[h]
\centering
\resizebox{.48\textwidth}{!}{
\begin{tabular}{|l|cc|cc|cc|}
\hline
\multicolumn{1}{|c|}{\multirow{2}{*}{\textbf{Attributes}}} & \multicolumn{2}{c|}{\textbf{AskUbuntu}}             & \multicolumn{2}{c|}{\textbf{Unix}}                  & \multicolumn{2}{c|}{\textbf{Serverfault}}           \\ \cline{2-7} 
\multicolumn{1}{|c|}{}                                     & \multicolumn{1}{c|}{\textbf{Train}} & \textbf{Test} & \multicolumn{1}{c|}{\textbf{Train}} & \textbf{Test} & \multicolumn{1}{c|}{\textbf{Train}} & \textbf{Test} \\ \hline
\textbf{Size}                                              & \multicolumn{1}{c|}{15,505}         & 203           & \multicolumn{1}{c|}{19,742}         & 241           & \multicolumn{1}{c|}{10908}          & 226           \\ \hline
\textbf{Year of questions}                                 & \multicolumn{1}{c|}{2019-20}        & 2021-23       & \multicolumn{1}{c|}{2019-20}        & 2021-23       & \multicolumn{1}{c|}{2019-20}        & 2021-23       \\ \hline
\textbf{Avg. length of questions}                          & \multicolumn{1}{c|}{254.38}         & 156.65        & \multicolumn{1}{c|}{205.85}         & 220.57        & \multicolumn{1}{c|}{259.81}         & 248.93        \\ \hline
\textbf{Avg. length of answers}                            & \multicolumn{1}{c|}{122.22}         & 217.16        & \multicolumn{1}{c|}{181.17}         & 210.02        & \multicolumn{1}{c|}{145.30}         & 161.73        \\ \hline
\end{tabular}
}
\caption{Dataset statistics.}
\label{tab:dataset}
\end{table}
\noindent In this experiment, we select three domain-specific datasets from open-source CQA platforms: AskUbuntu, ServerFault, and Unix, all of which originate from a low-resource domain with minimal properly annotated data available on these topics. These datasets, considered from June 2023, includes questions (title and body), a list of answers, an accepted answer flag, tags for the questions, and the posting dates and times for both questions and answers. For each question, the accepted answer serves as the ground truth. Due to the limited resources of the datasets and the high expenses associated with human involvement, we opt not to use human annotations. Further we apply several filtering procedures, such as duplicate question removal, non-specific answer removal, and length constraints (token limit in LLM) resulting in our dataset. Adopting the temporal splitting approach inspired by~\cite{HazraAGMC21,10.1007/978-3-031-26422-1_15}, we consider training set from 2019-2020 and test dataset from 2021-2023. Due to resource limitations, we randomly sample test dataset, the details of which are provided in Table~\ref{tab:dataset}. 
From the training set of each dataset, we construct an instruction-tuning dataset by pairing questions and answers in the format `$[INST] Question [\textbackslash INST] Answer: actual\_answer$'. We prepare this instruction data for each dataset.

\section{Methodology}
\label{sec:method}
In this section, we explain our proposed framework ~\textsc{GraphContextGen}. The overall framework is shown in Figure~\ref{fig2}. Our proposed framework consists of three modules -- (1) context retriever (2) KG-driven context enhancement (3) answer generation. Before explaining every module, we describe the problem in detail below.

\subsection{Preliminaries}
Given a community question answering (CQA) system, there is a collection of questions and their associated accepted answers, represented as $<Q,A_Q>$ = \{($q_1$, $a_{q_1}$), ($q_2$, $a_{q_2}$), ... ($q_n$, $a_{q_n}$)\}. The anchor question (query) is represented by $q$. We consider subset of question pool $Q_{pool}$, where $Q_{pool} \subset Q$. We represent our instruction tuned dataset as $\mathcal{D_{INST}}$ which contains instruction $\mathcal{INST}$, question pool $Q_{pool}$ and their accepted answer $A_{Q_{pool}}$. Our objective is formalized as follows.\\
\begin{algorithm}[!ht]
\tiny
\caption{\label{algo:algo1} \textsc{GraphContextGen}}
\begin{algorithmic}[1]
\State Input: Initial question pool $Q_{pool}$, query $q$, LLM $M$, instruction dataset $\mathcal{D_{INST}}$
 \Function{\textcolor{blue}{Retriever}}{$Q_{pool}, q$}
    \State Build $G(V, E)$, nodes $V = Q_{pool}$, edges $E \subseteq Q_{pool} \times Q_{pool}$ where $sim(q_a, q_b)> T$ $\forall$ $q_a, q_b \in Q_{pool}$
    \State Build extended graph $G^{'}(V^{'}, E^{'})$ where $V^{'} = V \cup {q}$ and $E^{'} = E \cup E_{q}$
    \State  $Q_{pool}^{ranked}$ = sort(~\emph{QueryAwarePageRank}($G^{'}$))
    \State Choose a set of top $k$ questions $Q_{top_{k}}^q$
\EndFunction
\Function{\textcolor{blue}{ ContextEnhancer}}{$Q_{top_{k}}^q, q$}
 \State context $C^q$ = $< Q_{top_{k}}^q, A_{Q_{top_{k}}^q}>$
 \State Extract $\tau^{init}(h, r, t)$ using LLM $M$ and REBEL from $C^q$ 
 \State $Ent(C^q)$ = ~\emph{EntitySetBuilder}($\tau^{init}$) where $Ent(C^q)$ contains set of entities ${ e_1, e_2, ..., e_n}$ 
 \State Extract triplets $\tau(h, r, t)$ from Wikidata for $h \in Ent(C^q)$
 \State Filtered triples set $\tau'$ if $t \in Ent(C^q)$ 
 \State Prepare triplets set $\tau^{f}$ =  $\tau^{init} \cup \tau'$
 \State Build sequence of sentences $S$ from all triplets $\tau^{f}$
 \State Enhanced context $C^q_{enc}$ =  $C_q \oplus S$ 
 \EndFunction
\Function{\textcolor{blue}{ AnswerGenerator}}{$C^q_{enc}, q$}
\State $M^{'}$ = SupervisedFineTuning($M, \mathcal{D_{INST}}$)
\State $a_{q}^{gen}$ = $M^{'}(C^q_{enc}, q)$
\EndFunction
\end{algorithmic}
\label{algo:GraphContextGen}
\end{algorithm}
\noindent\textbf{Context retriever}: In this module, we consider anchor question $q$ and the $Q_{pool}$ as input and output the most relevant questions from the $Q_{pool}$. The set of relevant questions is represented by  $Q^q_{top_{k}}$. We explain the working procedure of the module in subsequent sections.

\noindent \textbf{KG driven context enhancement}: 
This module takes the query $q$ and the final set of most relevant question $Q^q_{top_{k}}$ as input to formulate enhanced context. The initial context is represented by $C^q$ which is the $<Q^q_{top_{k}}$, $A_{Q^q_{top_{k}}}>$ pairs. Further, we represent the sequence of sentences by $S$ that are obtained from the entity extraction procedure and knowledge graph. Enhanced context is represented by \( C_{enc}^q \).\\
\noindent \textbf{Answer generation}: In this module, we provide the query $q$ and enhanced context \( C_{enc}^q \) as input to generate the answer denoted by \( a_{q}^{gen} \).
We denote the ground truth answer as \( a_{q}^{gt} \). We explain each of the above-mentioned steps in subsequent sections.

\subsection{Context retriever}
The objective of this module is to retrieve relevant previous questions given the query question. Our $RETRIEVER$ module in Algorithm~\ref{algo:GraphContextGen}, consists of two parts -- (I) question-question graph (Q-Q graph) construction and (II) retrieval of top relevant questions.

\noindent \textbf{(I) Q-Q graph construction}: 
We build a question-question graph (Q-Q graph) to obtain the relevant questions from the previously posted question pool $Q_{pool}$. In a Q-Q graph ($G(V, E)$), nodes ($V$) are the questions and the edges ($E$) are formed based on the cosine similarity between the  concatenated embeddings of the title and the body of two questions. We include the edge only if the similarity score crosses a particular threshold\footnote{Empirically identified based on graph density.}. 
The major motivation for building the Q-Q graph is that it can help to identify semantically similar questions based on the structural properties of the graph. This systematically prepared graph will be utilized to prioritize a set of existing questions given a query $q$. 

\noindent \textbf{(II) Retrieval of top relevant questions:}
For a given query $q$, we extend the existing Q-Q graph $G(V, E)$ to $G^{'}(V^{'}, E{'})$. We form the graph $G^{'}$ by including the query $q$ as a node and further measure the similarity with all the nodes in $G$. If the similarity score passes a threshold\footnote{We followed the same threshold used in Q-Q graph construction.}, the edges ($E_{q}$) are formed between question $q$ to the respective nodes in $G$ accordingly. 
We conceptualize that questions (in graph $G^{'}$) with high node centric score from the perspective of the query node $q$ could be considered as the relevant questions (nodes) to the query $q$. We use personalized PageRank (PPR)~\cite{10.1145/3394486.3403108} which introduces bias toward the query node and tailor the ranking based on the node preference (i.e., prior information). 

We obtain PPR scores for all the nodes (except $q$) in graph $G^{'}$. For the given query node $q$, we select the top $k$ relevant questions. This top $k$ question set is referred to as $Q^q_{top_{k}}$. 
We do the above mentioned process for all the queries in the query set.
 
\subsection{KG driven context enhancement}
From the previous module, we obtain $Q^q_{top_{k}}$ questions and their answers $A_{Q^q_{top_{k}}}$ and use them as context $C^q$ for a query $q$.
It is observed that LLMs lack in generating aligned answers for open ended questions~\cite{AI202380} even after providing the relevant context. In this module, we attempt to enhance the retrieved context $C^q$. In this process (see \textsc{ContextEnhancer} module in algorithm~\ref{algo:GraphContextGen}), we follow two major steps -- (i) entity identification and triplet formation, (ii) enhanced context formulation.\\    
\noindent \textbf{Entity identification and triplet formation}: In this stage, we first identify all the important information (e.g., entities) present in the $C^q$. For important information identification, 
we employ the LLM $M$ and REBEL~\cite{huguet-cabot-navigli-2021-rebel-relation} to obtain initial relation triplets ($\tau^{init}$) from the context $C^q$. We use a simple prompt plus the context $C^q$ to the LLM $M$ for relation triplet extraction task. In case of REBEL, we obtain the triplets by passing $C^q$ as input to their internal function. Note that a triplet consists of (head\_entity, tail\_entity, relation). We prepare a set $Ent(C^q)$ which contains all the entities present in these triplets.
 Further, we use Wikidata to obtain one hop neighbors of each entity and their relationship again in the form of triplets. We now consider all the new triplets ($\tau'$) as well as those in $\tau^{init}$ to prepare a new extended set of triplets $\tau^f$. We retain only those triplets in $\tau'$ whose head\_entity and tail\_entity are present in the original context $C^q$. 
\noindent \textbf{Enhanced context formulation}:  We construct a set of sub-contexts ($S$) in the form of sequence of sentences from triplet set $\tau^{f}$. Basically, we form the sentence by placing the head entity, the relation and the tail entity in sequence.
We finally construct the enhanced context $C_{enc}^q$ by concatenating the actual context $C_q$ and $S$. We illustrate the process in Figure~\ref{fig:enhCon}.

\subsection{Answer generation}
In this section, we use the enhanced context $C^q_{enc}$ and the given query $q$ to generate the answers using LLM. In this component, we use the LLM in two ways -- pretrained LLM and finetuned LLM. In the pretrained setup, we pass the enhanced context $C^q_{enc}$ and the query question $q$ as input and obtain the answer as output. In this setting, we use the LLM (model $M$) as black box. For fine tuned version, we utilize instruction dataset $\mathcal{D_{INST}}$ to efficiently fine tune the LLM $M$. The fine tuned model is represented as $M'$ Further, we use the enhanced context $C^q_{enc}$ and query $q$ as input to the fine tuned model $M'$ and obtain the generated answer $a^{gen}_{q}$.

\begin{table*}[]
\centering
\small
\resizebox{.99\textwidth}{!}{
\begin{tabular}{|l|ccccccccc|}
\hline
\multicolumn{1}{|c|}{\textbf{Method}}                & \multicolumn{3}{c|}{\textbf{AskUbuntu}}                                                                                  & \multicolumn{3}{c|}{\textbf{Unix}}                                                                                       & \multicolumn{3}{c|}{\textbf{ServerFault}}                                                           \\ \hline
\multicolumn{1}{|c|}{\multirow{2}{*}{\textbf{Size \big\uparrow}}} & \multicolumn{1}{c|}{\textbf{BERTScore}} & \multicolumn{1}{c|}{\textbf{ROUGE 1}} & \multicolumn{1}{c|}{\textbf{ROUGE L}} & \multicolumn{1}{c|}{\textbf{BERTScore}} & \multicolumn{1}{c|}{\textbf{ROUGE 1}} & \multicolumn{1}{c|}{\textbf{ROUGE L}} & \multicolumn{1}{c|}{\textbf{BERTScore}} & \multicolumn{1}{c|}{\textbf{ROUGE 1}} & \textbf{ROUGE L} \\ \cline{2-10} 
\multicolumn{1}{|c|}{}                               & \multicolumn{9}{c|}{\textbf{macro-F1 score}}                                                                                                                                                                                                                                                                                                              \\ \hline
\textbf{Phi (1.5B)~\cite{li2023textbooks}}                                  & \multicolumn{1}{c|}{\cellcolor{cyan!10}0.803}               & \multicolumn{1}{c|}{\cellcolor{magenta!10}0.219}            & \multicolumn{1}{c|}{\cellcolor{magenta!10}0.202}            & \multicolumn{1}{c|}{0.792}               & \multicolumn{1}{c|}{\cellcolor{magenta!30}0.191}            & \multicolumn{1}{c|}{\cellcolor{magenta!50}0.176}            & \multicolumn{1}{c|}{0.790}               & \multicolumn{1}{c|}{\cellcolor{magenta!50}0.202}            & \cellcolor{magenta!50}0.183            \\ \hline
\textbf{Falcon (7B)~\cite{falcon}}                                 & \multicolumn{1}{c|}{0.718}               & \multicolumn{1}{c|}{0.167}            & \multicolumn{1}{c|}{0.153}            & \multicolumn{1}{c|}{\cellcolor{cyan!10}0.794}               & \multicolumn{1}{c|}{0.151}            & \multicolumn{1}{c|}{0.138}            & \multicolumn{1}{c|}{0.801}                    & \multicolumn{1}{c|}{0.181}                 &   {0.166}               \\ \hline
\textbf{MPT (7B)~\cite{MosaicML2023Introducing}}                                    & \multicolumn{1}{c|}{0.738}               & \multicolumn{1}{c|}{0.156}            & \multicolumn{1}{c|}{0.147}            & \multicolumn{1}{c|}{0.786}               & \multicolumn{1}{c|}{0.138}            & \multicolumn{1}{c|}{0.127}            & \multicolumn{1}{c|}{0.709}                    & \multicolumn{1}{c|}{0.144}                 &   {0.132}               \\ \hline
\textbf{StackLlama (7B)~\cite{beeching2023stackllama}}                             & \multicolumn{1}{c|}{0.797}               & \multicolumn{1}{c|}{0.136}            & \multicolumn{1}{c|}{0.130}            & \multicolumn{1}{c|}{0.774}               & \multicolumn{1}{c|}{0.122}            & \multicolumn{1}{c|}{0.112}            & \multicolumn{1}{c|}{0.785}                    & \multicolumn{1}{c|}{0.131}                 &  {0.120}                \\ \hline
\textbf{Llama2 (7B)~\cite{touvron2023llama}}                                 & \multicolumn{1}{c|}{\cellcolor{cyan!30}0.809}               & \multicolumn{1}{c|}{\cellcolor{magenta!30}0.221}            & \multicolumn{1}{c|}{\cellcolor{magenta!30}0.204}            & \multicolumn{1}{c|}{0.792}               & \multicolumn{1}{c|}{0.178}            & \multicolumn{1}{c|}{0.163}            & \multicolumn{1}{c|}{\cellcolor{cyan!10}0.813}               & \multicolumn{1}{c|}{0.183}            & 0.167            \\ \hline
\textbf{Flan-t5-xxl (11B)~\cite{roberts2022t5x}}                           & \multicolumn{1}{c|}{0.795}               & \multicolumn{1}{c|}{0.131}                 & \multicolumn{1}{c|}{0.114}                 & \multicolumn{1}{c|}{0.792}               & \multicolumn{1}{c|}{0.106}            & \multicolumn{1}{c|}{0.098}            & \multicolumn{1}{c|}{\cellcolor{cyan!30}0.816}               & \multicolumn{1}{c|}{0.135}            & 0.124            \\ \hline
\textbf{Vicuna (13B)~\cite{vicuna2023}}                                & \multicolumn{1}{c|}{0.741}               & \multicolumn{1}{c|}{0.177}            & \multicolumn{1}{c|}{0.163}            & \multicolumn{1}{c|}{0.789}               & \multicolumn{1}{c|}{\cellcolor{magenta!10}0.181}            & \multicolumn{1}{c|}{0.165}            & \multicolumn{1}{c|}{0.810}               & \multicolumn{1}{c|}{0.185}            & 0.169            \\ \hline
\textbf{Llama2 (13B)~\cite{touvron2023llama}}                                & \multicolumn{1}{c|}{\cellcolor{cyan!70}0.810}               & \multicolumn{1}{c|}{\cellcolor{magenta!50}0.227}            & \multicolumn{1}{c|}{\cellcolor{magenta!50}0.211}            & \multicolumn{1}{c|}{\cellcolor{cyan!70}0.806}               & \multicolumn{1}{c|}{\cellcolor{magenta!50}0.191}            & \multicolumn{1}{c|}{\cellcolor{magenta!30}0.175}            & \multicolumn{1}{c|}{\cellcolor{cyan!70}0.819}               & \multicolumn{1}{c|}{\cellcolor{magenta!30}0.189}            & \cellcolor{magenta!30}0.176            \\ \hline
\textbf{Gpt-neox (20B)~\cite{black-etal-2022-gpt}}                              & \multicolumn{1}{c|}{0.724}               & \multicolumn{1}{c|}{0.136}            & \multicolumn{1}{c|}{0.127}            & \multicolumn{1}{c|}{0.776}               & \multicolumn{1}{c|}{0.140}            & \multicolumn{1}{c|}{0.131}            & \multicolumn{1}{c|}{0.799}                    & \multicolumn{1}{c|}{0.152}                 &  {0.161}                \\ \hline
\textbf{Falcon (40B)~\cite{falcon}}                                & \multicolumn{1}{c|}{0.721}               & \multicolumn{1}{c|}{0.182}            & \multicolumn{1}{c|}{0.167}            & \multicolumn{1}{c|}{\cellcolor{cyan!30}0.801}                    & \multicolumn{1}{c|}{0.179}                 & \multicolumn{1}{c|}{\cellcolor{magenta!10}0.171}                 & \multicolumn{1}{c|}{0.812}                    & \multicolumn{1}{c|}{\cellcolor{magenta!10}0.186}                 &      {\cellcolor{magenta!10}0.173}            \\ \hline
\end{tabular}
}
\caption{Comparison of zero-shot learning performance of various models of different sizes across the AskUbuntu, Unix, and ServerFault datasets. Metrics include BERTScore, ROUGE 1, and ROUGE L scores. The cell color intensity indicates the relative performance, with darker shades representing higher values. The best results are marked with the darkest shade of \colorbox{cyan!70}{cyan} for BERTScore \& \colorbox{magenta!50}{magenta} for ROUGE scores.}
\label{tab:zeroshot}
\end{table*}

\begin{table*}[]
\centering
\small
\resizebox{.99\textwidth}{!}{
\begin{tabular}{|lccclccclcccl|}
\hline
\multicolumn{1}{|c|}{\multirow{3}{*}{\textbf{Method}}}                                                                      & \multicolumn{4}{c|}{\textbf{AskUbuntu}}                                                                                                                             & \multicolumn{4}{c|}{\textbf{Unix}}                                                                                                                                  & \multicolumn{4}{c|}{\textbf{ServerFault}}                                                                                                      \\ \cline{2-13} 
\multicolumn{1}{|c|}{}                                                                                     & \multicolumn{1}{c|}{\textbf{BERTScore}} & \multicolumn{1}{c|}{\textbf{ROUGE 1}} & \multicolumn{1}{c|}{\textbf{ROUGE L}} & \multicolumn{1}{l|}{\textbf{FactSumm}} & \multicolumn{1}{c|}{\textbf{BERTScore}} & \multicolumn{1}{c|}{\textbf{ROUGE 1}} & \multicolumn{1}{c|}{\textbf{ROUGE L}} & \multicolumn{1}{l|}{\textbf{FactSumm}} & \multicolumn{1}{c|}{\textbf{BERTScore}} & \multicolumn{1}{c|}{\textbf{ROUGE 1}} & \multicolumn{1}{c|}{\textbf{ROUGE L}} & \textbf{FactSumm} \\ \cline{2-13} 
\multicolumn{1}{|c|}{\textbf{}}                                                                            & \multicolumn{3}{c|}{\textbf{macro-F1 score}}                                                                             & \multicolumn{1}{l|}{\textbf{}}           & \multicolumn{3}{c|}{\textbf{macro-F1 score}}                                                                             & \multicolumn{1}{l|}{\textbf{}}           & \multicolumn{3}{c|}{\textbf{macro-F1 score}}                                                                             & \textbf{}           \\ \hline
\multicolumn{13}{|c|}{\textbf{Pre-LLM era}}                                                                                                                                                                                                                                                                                                                                                                                                                                                                                                                                                             \\ \hline
\multicolumn{1}{|l|}{\textbf{AnswerBot {\cite{Xu:2017}}}}                                                   & \multicolumn{1}{c|}{0.803}               & \multicolumn{1}{c|}{\cellcolor{magenta!50}0.236}            & \multicolumn{1}{c|}{0.111}            & \multicolumn{1}{c|}{0.578}                    & \multicolumn{1}{c|}{0.791}               & \multicolumn{1}{c|}{\cellcolor{magenta!50}0.191}            & \multicolumn{1}{c|}{0.091}            & \multicolumn{1}{c|}{0.583}                    & \multicolumn{1}{c|}{0.802}               & \multicolumn{1}{c|}{0.191}            & \multicolumn{1}{c|}{0.094}            &      \multicolumn{1}{c|}{0.642}               \\ \hline
\multicolumn{1}{|l|}{\textbf{GenQA {\cite{hsu:2021}}}}                                                      & \multicolumn{1}{c|}{0.781}                    & \multicolumn{1}{c|}{0.095}                 & \multicolumn{1}{c|}{0.071}                 & \multicolumn{1}{c|}{0.551}                    & \multicolumn{1}{c|}{0.55}                    & \multicolumn{1}{c|}{0.048}                 & \multicolumn{1}{c|}{0.04}                 & \multicolumn{1}{c|}{0.427}                    & \multicolumn{1}{c|}{0.668}                    & \multicolumn{1}{c|}{0.059}                 & \multicolumn{1}{c|}{0.045}                 &              \multicolumn{1}{c|}{0.662}       \\ \hline
\multicolumn{1}{|l|}{\textbf{TechSumBot {~\cite{10.1145/3551349.3560421}}}}                                             & \multicolumn{1}{c|}{0.781}                    & \multicolumn{1}{c|}{0.100}                 & \multicolumn{1}{c|}{0.05}                 & \multicolumn{1}{c|}{0.580}                    & \multicolumn{1}{c|}{0.776}                    & \multicolumn{1}{c|}{0.077}                 & \multicolumn{1}{c|}{0.039}                 & \multicolumn{1}{c|}{0.563}                    & \multicolumn{1}{c|}{0.781}                    & \multicolumn{1}{c|}{0.064}                 & \multicolumn{1}{c|}{0.034}                 &      \multicolumn{1}{c|}{0.655}               \\ \hline
\multicolumn{13}{|c|}{\textbf{LLM era (best performing LLM from Table~\ref{tab:zeroshot} is used, i.e., Llama2 (13B))}}                                                                                                                                                                                                                                                                                                                                                                                                                                                                                                                                                            \\ \hline
\multicolumn{1}{|l|}{\textbf{[w/o INST]~\textsc{TextGen}}}                                                          & \multicolumn{1}{c|}{0.812}               & \multicolumn{1}{c|}{0.217}            & \multicolumn{1}{c|}{\cellcolor{magenta!10}0.202}            & \multicolumn{1}{c|}{0.612}                    & \multicolumn{1}{c|}{0.809}                    & \multicolumn{1}{c|}{0.179}                 & \multicolumn{1}{c|}{0.162}                 & \multicolumn{1}{c|}{0.683}                    & \multicolumn{1}{c|}{0.810}                    & \multicolumn{1}{c|}{0.179}                 & \multicolumn{1}{c|}{0.166}                 & \multicolumn{1}{c|}{0.733}                     \\ \hline
\multicolumn{1}{|l|}{\textbf{[w/o INST]~\textsc{TextContextGen}}}                                                          & \multicolumn{1}{c|}{0.827}               & \multicolumn{1}{c|}{\cellcolor{magenta!30}0.223}            & \multicolumn{1}{c|}{\cellcolor{magenta!30}0.204}            & \multicolumn{1}{c|}{0.619}                    & \multicolumn{1}{c|}{0.818}                    & \multicolumn{1}{c|}{0.184}                 & \multicolumn{1}{c|}{0.168}                 & \multicolumn{1}{c|}{0.683}                    & \multicolumn{1}{c|}{0.823}                    & \multicolumn{1}{c|}{0.198}                 & \multicolumn{1}{c|}{0.175}                 & \multicolumn{1}{c|}{0.738}                    \\ \hline
\multicolumn{1}{|l|}{\textbf{[w/o INST]~\textsc{GraphGen}}}                                                          & \multicolumn{1}{c|}{0.823}               & \multicolumn{1}{c|}{0.204}            & \multicolumn{1}{c|}{0.188}            & \multicolumn{1}{c|}{0.619}                    & \multicolumn{1}{c|}{0.809}                    & \multicolumn{1}{c|}{0.181}                 & \multicolumn{1}{c|}{0.162}                 & \multicolumn{1}{c|}{0.683}                    & \multicolumn{1}{c|}{0.816}                    & \multicolumn{1}{c|}{0.182}                 & \multicolumn{1}{c|}{0.166}                 &  \multicolumn{1}{c|}{0.737}                   \\ \hline
\multicolumn{1}{|l|}{\textbf{[w/o INST]~\textsc{GraphContextGen}}}                                                          & \multicolumn{1}{c|}{\cellcolor{cyan!10}0.831}               & \multicolumn{1}{c|}{\cellcolor{magenta!10}0.222}            & \multicolumn{1}{c|}{\cellcolor{magenta!50}0.206}            & \multicolumn{1}{c|}{0.621}                    & \multicolumn{1}{c|}{0.822}                    & \multicolumn{1}{c|}{0.184}                 & \multicolumn{1}{c|}{\cellcolor{magenta!10}0.169}                 & \multicolumn{1}{c|}{0.685}                    & \multicolumn{1}{c|}{0.823}                    & \multicolumn{1}{c|}{0.197}                 & \multicolumn{1}{c|}{0.175}                 & \multicolumn{1}{c|}{\cellcolor{blue!10}0.738}                    \\ \hline
\multicolumn{1}{|l|}{\textbf{FineTuned~\textsc{Gen} Zero-Shot}} & \multicolumn{1}{c|}{0.815}               & \multicolumn{1}{c|}{0.203}            & \multicolumn{1}{c|}{0.187}            & \multicolumn{1}{c|}{0.608}                    & \multicolumn{1}{c|}{0.812}               & \multicolumn{1}{c|}{0.183}            & \multicolumn{1}{c|}{0.167}            & \multicolumn{1}{c|}{0.661}                    & \multicolumn{1}{c|}{0.821}               & \multicolumn{1}{c|}{0.195}            & \multicolumn{1}{c|}{0.179}            &      \multicolumn{1}{c|}{0.733}               \\ \hline
\multicolumn{1}{|l|}{\textbf{\textsc{TextGen}}} & \multicolumn{1}{c|}{0.821}               & \multicolumn{1}{c|}{0.183}            & \multicolumn{1}{c|}{0.170}            & \multicolumn{1}{c|}{0.623}                    & \multicolumn{1}{c|}{\cellcolor{cyan!10}0.823}               & \multicolumn{1}{c|}{\cellcolor{magenta!10}0.186}            & \multicolumn{1}{c|}{\cellcolor{magenta!30}0.169}            & \multicolumn{1}{c|}{0.684}                    & \multicolumn{1}{c|}{0.829}               & \multicolumn{1}{c|}{0.197}            & \multicolumn{1}{c|}{0.179}            &      \multicolumn{1}{c|}{\cellcolor{blue!20}0.738}               \\ \hline
\multicolumn{1}{|l|}{\textbf{\textsc{TextContextGen}}} & \multicolumn{1}{c|}{\cellcolor{cyan!30}0.833}               & \multicolumn{1}{c|}{0.221}            & \multicolumn{1}{c|}{0.200}            & \multicolumn{1}{c|}{\cellcolor{blue!10}0.636}                    & \multicolumn{1}{c|}{\cellcolor{cyan!30}0.834}               & \multicolumn{1}{c|}{0.182}            & \multicolumn{1}{c|}{0.161}            & \multicolumn{1}{c|}{\cellcolor{blue!10}0.689}                    & \multicolumn{1}{c|}{\cellcolor{cyan!10}0.831}               & \multicolumn{1}{c|}{\cellcolor{magenta!10}0.198}            & \multicolumn{1}{c|}{\cellcolor{magenta!10}0.180}            &       \multicolumn{1}{c|}{\cellcolor{blue!40}0.739}              \\ \hline
\multicolumn{1}{|l|}{\textbf{\textsc{GraphGen}}} & \multicolumn{1}{c|}{0.827}               & \multicolumn{1}{c|}{0.182}            & \multicolumn{1}{c|}{0.170}            & \multicolumn{1}{c|}{\cellcolor{blue!20}0.636}                    & \multicolumn{1}{c|}{0.817}               & \multicolumn{1}{c|}{0.183}            & \multicolumn{1}{c|}{0.164}            & \multicolumn{1}{c|}{\cellcolor{blue!20}0.691}                    & \multicolumn{1}{c|}{\cellcolor{cyan!30}0.831}               & \multicolumn{1}{c|}{\cellcolor{magenta!30}0.198}            & \multicolumn{1}{c|}{\cellcolor{magenta!30}0.180}            &      \multicolumn{1}{c|}{0.737}               \\ \hline
\multicolumn{1}{|l|}{\textbf{\textsc{GraphContextGen*}}}                                                             & \multicolumn{1}{c|}{\cellcolor{cyan!70}0.840}               & \multicolumn{1}{c|}{0.214}            & \multicolumn{1}{c|}{0.189}            & \multicolumn{1}{c|}{\cellcolor{blue!40}0.639}                    & \multicolumn{1}{c|}{\cellcolor{cyan!70}0.837}                    & \multicolumn{1}{c|}{\cellcolor{magenta!30}0.187}                 & \multicolumn{1}{c|}{\cellcolor{magenta!50}0.169}                 & \multicolumn{1}{c|}{\cellcolor{blue!40}0.693}                    & \multicolumn{1}{c|}{\cellcolor{cyan!70}0.839}                    & \multicolumn{1}{c|}{\cellcolor{magenta!50}0.198}                 & \multicolumn{1}{c|}{\cellcolor{magenta!50}0.181}                 &        \multicolumn{1}{c|}{0.737}             \\ \hline
\end{tabular}
}
\caption{Comparison of various question-answering and summarization methods on AskUbuntu, Unix, and ServerFault platforms using evaluation metrics BERTScore, ROUGE 1, ROUGE L, and FactSumm. Methods are categorized into those developed before the LLM era (pre-LLM era) and those developed during the LLM era. * $p$-value $< 0.05$ on comparison with pre-LLM era models. The best results are marked with the darkest shade of \colorbox{cyan!70}{cyan} for BERTScore, \colorbox{magenta!50}{magenta} for ROUGE score \& \colorbox{blue!40}{blue} for FactSumm.} 
\label{tab:maintable}
\end{table*}
   
\section{Experimental setup}
\label{sec:expsetup}
\noindent\textbf{Baselines}: In this work, we use various methods as baselines. Some of the baselines are proposed by us which we believe are very competitive to our best approach.

\noindent \textbf{Pre-LLM era baselines}: We compare our approach with SOTA answer generation/summarization works such as \textbf{AnswerBot}~\cite{Xu:2017, Cai:2019}, \textbf{GenQA}~\cite{hsu:2021} and \textbf{TechSumBot}~\cite{10172591} (see Appendix~\ref{sec:prellmbase}). Due to unavailability of the codebase and unclear implementation details, we could not compare this paper~\cite{deng2019joint} with our method.
\noindent \textbf{Zeroshot LLMs}: In this setting, we use various competitive LLMs to generate the answer of the given question. The LLMs are of different parameter sizes (7B to 40B). Such a choice enables us to understand how well models with diverse parameter sizes perform in zero shot setting.
\noindent \textbf{[w/o INST]~\textsc{TextGen}}: In this setup, we use a vector database (chromaDB\footnote{https://docs.trychroma.com/getting-started} and FAISS\footnote{https://python.langchain.com/docs/integrations/vectorstores/faiss}) containing all the training set questions. We compute contextual similarity between query $q$ and all the questions in database. We rank the questions in database based on the cosine similarity scores (higher scores get top ranks) and retrieve top $k$ questions. Further we use the top $k$ questions and their actual answers as few shot examples to the pretrained LLM for generating the answer. 
\noindent \textbf{[w/o INST]~\textsc{TextContextGen}}: In this setup, we retrieve the top $k$ questions and their answers using the same method as ~\textbf{[w/o INST]~\textsc{TextGen}}. Subsequently, we use context enhancement component of our approach to enhance the context. Further we provide the enhanced context and the query as input to the pretrained LLM and obtain the generated answer.
\noindent \textbf{[w/o INST]~\textsc{GraphGen}}: In this setup, we use the \textsc{Retriever} module of our algorithm to retrieve the top $k$ questions. We use $k$ questions and their answers as few shot examples to the pretrained LLM for generating the answer. 
\noindent \textbf{[w/o INST]~\textsc{GraphContextGen}}: We follow the retrieval step from ~\textbf{[w/o INST]~\textsc{GraphGen}}. Further we use our \textsc{ContextEnhancer} module to enhance the context. 
\noindent \textbf{\textsc{FineTuned Gen Zero-Shot}}: In this setting, we use instruction fine tuned LLM in zero shot settings. Here, we pass the questions as input and the fine tuned LLM generates the answer.
\noindent \textbf{\textsc{TextGen}}: This setup is same as \textbf{[w/o INST]~\textsc{TextGen}}. However, we use our instruction fine tuned LLM for generation.
\noindent \textbf{\textsc{GraphGen}}: This setup is same as \textbf{[w/o INST]~\textsc{GraphGen}}. However, we use our instruction fine tuned LLM for generation.

\noindent \textbf{Parameter setting\footnote{Values of all these hyperparameters are obtained through grid search.}}: In our method ~\textsc{GraphContextGen}, we use Flag embedding~\cite{bgeEmbedding} (bge-large-en) to obtain embedding for each question in the training set. The dimension of the embedding is 1024. We construct the edges of the Q-Q graph if the embedding cosine similarity between two questions cross a threshold of 0.8~\footnote{Empirically computed based on graph density.}. In PPR algorithm, the $\alpha$ value is set to 0.85, $max\_iter$ is set to 100 and $tol$ is set to 1e-6. The $k$ value is set to 2~\footnote{Empirically identified to fit the whole context within the acceptable token limit of the LLMs.}. For parameter settings of instruction tuned models, see Appendix ~\ref{sec:insttunehyp}.\\ 
\noindent \textbf{Evaluation metrics}: We have used three metrics -- ROUGE score\footnote{https://huggingface.co/spaces/evaluate-metric/rouge}, BERT score~\cite{bert-score, zhang2020bertscore} and FactSumm score~\cite{factsumm} for automatic evaluation of generated answers. Note that the FactSumm~\cite{factsumm} package extracts the facts from the generated text and the ground truth text and computes an overall score based on the fact overlap and fact mismatch. This package has also been used in earlier works~\cite{Liu2021CO2SumCL, qian2023webbrain} to measure the factual accuracy of the generated text. 
\section{Results}
\label{sec:results}
The Table \ref{tab:zeroshot} notes the BERTScore, ROUGE 1, ROUGE L (macro-F1) values achieved by different LLMs in a zero-shot setup across three domains -- AskUbuntu, Unix, and ServerFault.

\noindent\textbf{Zero-shot answer generation by different LLMs}: For AskUbuntu, Llama2 (13B) achieves the highest BERTScore (0.810), ROUGE 1 (0.227), and ROUGE L (0.211). For Unix, Llama2 (13B) leads in BERTScore (0.806), while Vicuna (13B) tops ROUGE 1 (0.181), and Llama2 (13B) tops ROUGE L (0.175). For ServerFault, Llama2 (13B) dominates BERTScore (0.819), Vicuna (13B) leads in ROUGE 1 (0.185), and Llama2 (13B) in ROUGE L (0.176). Performances are not always proportional to model size, as Llama2 (13B) often outperforms larger models like Gpt-neox (20B) and Falcon (40B). Models of similar sizes also display varied performances, indicating the importance of architecture and training methods.

\noindent\textbf{Main results}: Table \ref{tab:maintable} compares baseline results with our proposed method. \textbf{Pre-LLM era:} AnswerBot shows competitive BERTScore (0.803, 0.791, 0.802) for AskUbuntu, Unix, and ServerFault, respectively, while GenQA underperforms on Unix (BERTScore 0.55). AnswerBot achieves the highest FactSumm score across all platforms. \textbf{LLM era:} Llama2 (13B) is the reference model for generating answers. Our model, \textsc{GraphContextGen}, outperforms all baselines in BERTScore and FactSumm for AskUbuntu and Unix, producing factually more correct answers. For ROUGE 1 and ROUGE L, \textsc{GraphContextGen} shows competitive performance in Unix and ServerFault. Models from the LLM era generally outperform pre-LLM models in these metrics.

\begin{figure}[htbp]
    \centering
    \includegraphics[width=0.48\textwidth]{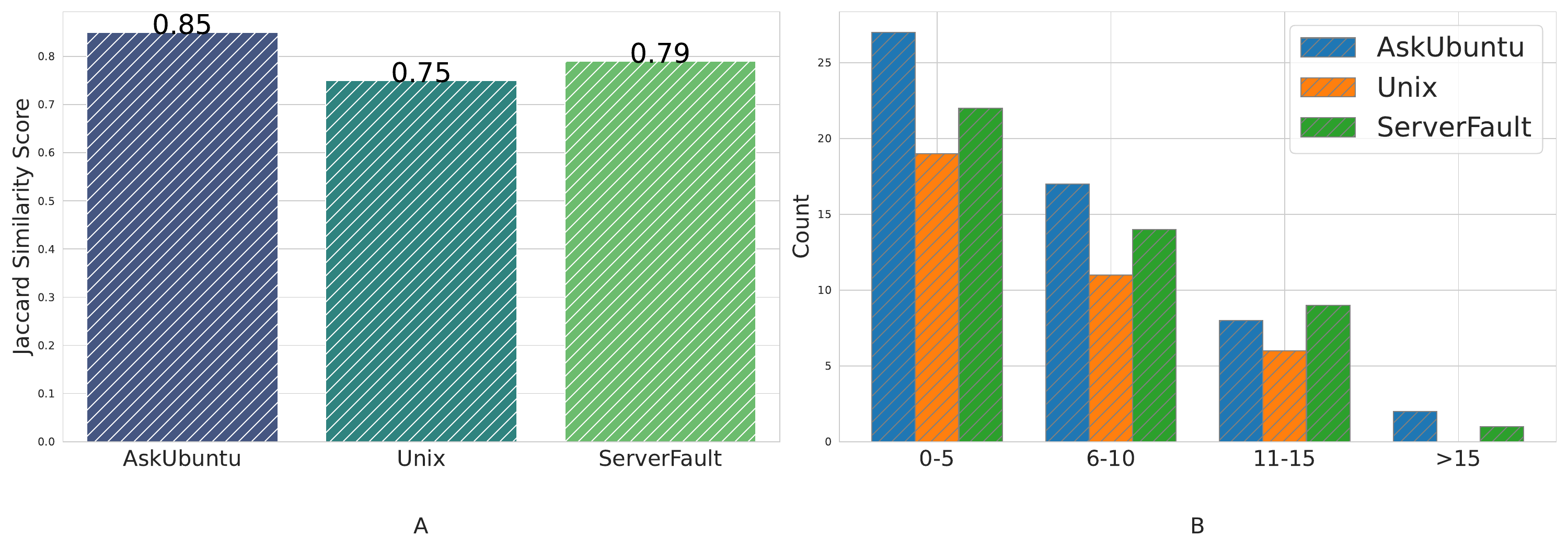} 
    \caption{Performance metrics on community datasets using the UniNER model. (A) Jaccard similarity scores illustrate the level of overlap between predicted entities and actual entities. (B) Triplet overlap distribution across different ranges, provide insights into the depth of entity matching in the model's predictions.}
    \label{fig:my_label}
\end{figure}

\noindent\textbf{Grounding of the generated answers}: We use UniversalNER~\cite{zhou2023universalner} to identify entities in the ground truth and generated answers. The Jaccard similarity between entity sets for our model is 0.85, 0.75, and 0.79 for AskUbuntu, Unix, and ServerFault, respectively (Figure~\ref{fig:my_label}(A)). The overlap in the number of triplets is shown in Figure~\ref{fig:my_label}(B), indicating our model's answers are rich in entities and relationships present in the ground truth.
\if{0}\noindent\textbf{Ablation study}: In this section, we attempt to understand how well each component is working and contributing to the overall performance. Here, we have done this study mainly from three different angles. \am{Where are these results?}
\vspace*{-0.1cm}
\begin{enumerate}[(i)]
    \item \textit{Based on embedding algorithms}: The performance differences between sentence BERT~\cite{reimers2019sentencebert} and BGE embeddings~\cite{bgeEmbedding} within the \textsc{TextGen} model are evident. The BGE variant demonstrates a slightly improved performance over sentence BERT.
    \item \textbf{Based on one-shot}: When focusing on one-shot methodologies, the \textsc{GraphGen} model achieves marginally better scores compared to the \textsc{TextGen} model. This suggests that the incorporation of graph structures potentially aids in better knowledge retrieval
    \item \textbf{Based on Only Knowledge Graph Context:} Focusing solely on the knowledge graph context, two variants emerge: \textsc{OnlyContextGen}[FT] (From Text) and \textsc{OnlyContextGen}[FG] (From Graph). The ``From Graph" variant consistently exhibits slightly higher scores across all platforms than its ``From Text" counterpart. This underscores the importance of leveraging structured graph data for enhanced performance in our application.
\end{enumerate}
\vspace*{-0.1cm}\fi

\begin{figure}[h]
    \centering
    \includegraphics[width=1.0\columnwidth]{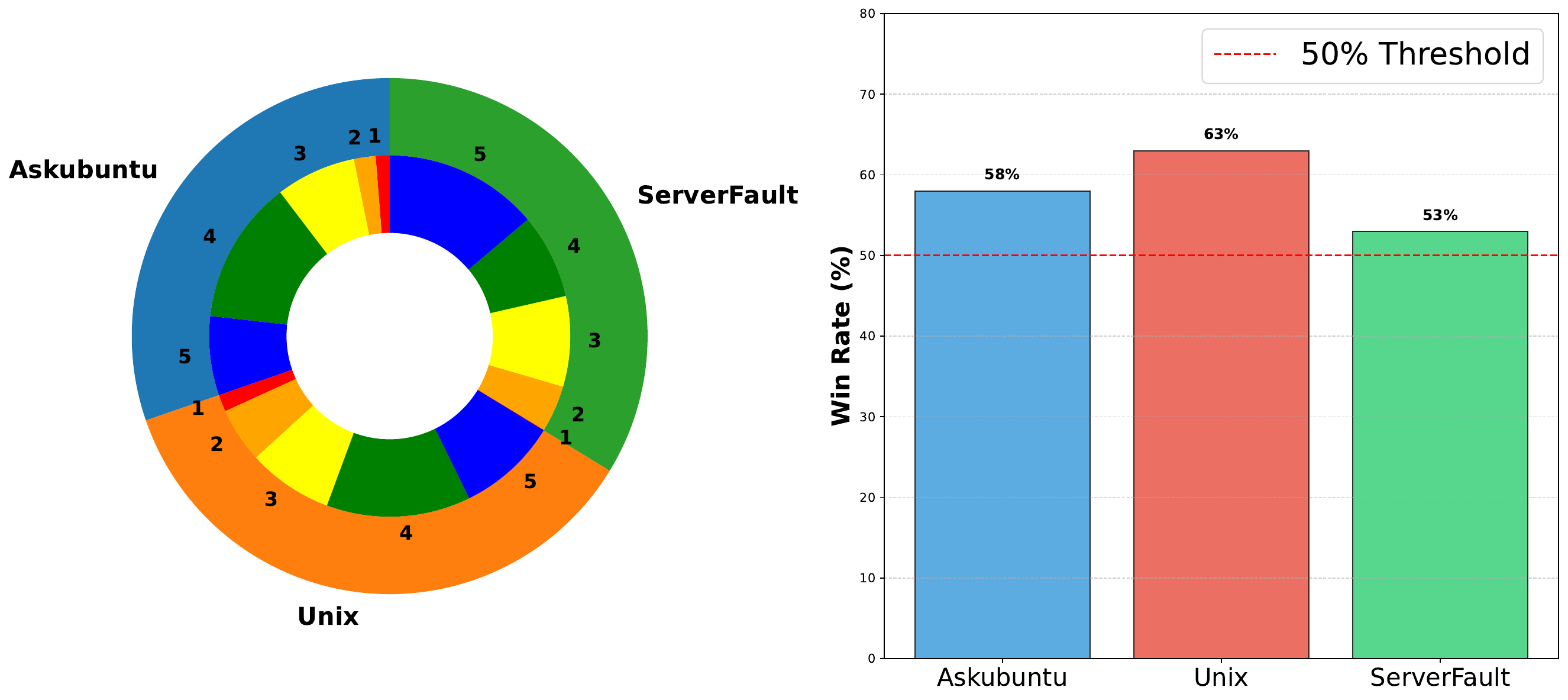}
    \caption{Comparative analysis of human feedback on the generated answers for test questions drawn from Askubuntu, Unix, and ServerFault. On the left, a dual-layer pie chart breaks down the total number of answers and their respective ratings from 1 to 5. The right side displays a bar graph indicating the percentage of wins for answers from each community, benchmarked against a 50\% threshold. Notably, a majority of the ratings lean toward the higher end, indicating overall positive reception.}
    \label{fig:humanfeedback}
    \vspace*{-0.3cm}
\end{figure}


\subsection{Ablation study}
In this section, we attempt to understand how well each component of our model is working and contributing to the overall performance. Here, we have done this study mainly from three different angles (see Table~\ref{tab:ablation}).
\begin{compactitem}
    \item \textit{Based on embedding algorithms}: The performance differences between sentence BERT~\cite{reimers2019sentencebert} and BGE embeddings~\cite{bgeEmbedding} within the \textsc{TextGen} model are evident. The BGE variant demonstrates a slightly improved performance over sentence BERT.
    \item \textit{Based on one-shot}: When focusing on one-shot methodologies, the \textsc{GraphGen} model achieves marginally better scores compared to the \textsc{TextGen} model. This suggests that the incorporation of graph structures potentially aids in better knowledge retrieval
    \item \textit{Based on only knowledge graph context}: Focusing solely on the knowledge graph context, two variants emerge: \textsc{OnlyContextGen}[FT] (from text) and \textsc{OnlyContextGen}[FG] (from graph). The `From Graph' variant consistently exhibits slightly higher scores across all platforms than its `From Text' counterpart. This underscores the importance of leveraging structured graph data for enhanced performance in our application.
\end{compactitem}

\begin{table}[]
\small
\scalebox{0.67}{
\begin{tabular}{|llll|}
\hline
\multicolumn{1}{|c|}{\textbf{Method}}                                                                       & \multicolumn{1}{l|}{\textbf{AskUbuntu}} & \multicolumn{1}{l|}{\textbf{Unix}} & \textbf{ServerFault} \\ \hline
\multicolumn{4}{|c|}{\textit{\textbf{Based on embedding algorithms}}}                                                                                                                                      \\ \hline
\multicolumn{1}{|l|}{\textbf{\begin{tabular}[c]{@{}l@{}}[w/o INST]~\textsc{TextGen} (Sentence BERT)\end{tabular}}}         & \multicolumn{1}{c|}{0.808}                   & \multicolumn{1}{c|}{0.783}              &      \multicolumn{1}{c|}{0.803}                \\ \hline
\multicolumn{1}{|l|}{\textbf{\begin{tabular}[c]{@{}l@{}}[w/o INST]~\textsc{TextGen} (BGE)\end{tabular}}}       & \multicolumn{1}{c|}{0.812}                   & \multicolumn{1}{c|}{0.809}              &    \multicolumn{1}{c|}{0.810}                  \\ \hline
\multicolumn{4}{|c|}{\textit{\textbf{Based on one-shot}}}                                                                                                                                      \\ \hline
\multicolumn{1}{|l|}{\textbf{\begin{tabular}[c]{@{}l@{}}[w/o INST]~\textsc{TextGen}\end{tabular}}}         & \multicolumn{1}{c|}{0.809}                   & \multicolumn{1}{c|}{0.807}              &                \multicolumn{1}{c|}{0.807}      \\ \hline
\multicolumn{1}{|l|}{\textbf{\begin{tabular}[c]{@{}l@{}}[w/o INST]~\textsc{GraphGen}\end{tabular}}}       & \multicolumn{1}{c|}{0.818}                   & \multicolumn{1}{c|}{0.809}              &    \multicolumn{1}{c|}{0.812}                  \\ \hline
\multicolumn{4}{|c|}{\textit{\textbf{Based on only knowledge graph context}}}                                                                                                                                      \\ \hline
\multicolumn{1}{|l|}{\textbf{\begin{tabular}[c]{@{}l@{}}[w/o INST]~\textsc{OnlyContextGen}[FT]\end{tabular}}}         & \multicolumn{1}{c|}{0.819}                   & \multicolumn{1}{c|}{0.807}              &   \multicolumn{1}{c|}{0.819}                   \\ \hline
\multicolumn{1}{|l|}{\textbf{\begin{tabular}[c]{@{}l@{}}[w/o INST]~\textsc{OnlyContextGen}[FG]\end{tabular}}}       & \multicolumn{1}{c|}{0.827}                   & \multicolumn{1}{c|}{0.820}              &    \multicolumn{1}{c|}{0.818}                  \\ \hline
\end{tabular}
}
\caption{Ablation study highlighting the performance of various methods. FT: From text, FG: From graph.}
\label{tab:ablation}
\end{table}

\section{Human evaluation}
In the results of the automatic evaluation, it is essential to remember that low values of BERTScore or FactSumm might also correspond to lower-quality ground truth answers posted by humans. Thus, there is a possibility that the model-generated answers are superior in quality compared to the ground truth answers. Such points can be verified only by human judgment experiments presented in Figure~\ref{fig:humanfeedback} and the Appendix~\ref{appd:humanannotation}.

\section{Error Analysis}
This section presents a systematic error analysis highlighting the error types and corresponding examples (see Table~\ref{tab:erroranalysis}).

\noindent\textit{\textbf{Misaligned retrieval outcomes}}: This misalignment occurs when the retrieved content, accurate in its context, doesn't match the user's intended query, leading to off-target responses due to the generation's reliance on the retrieved data. On platforms like AskUbuntu, Unix, and ServerFault, overlapping themes, like a `boot issues' query retrieving `USB booting' content instead of `system booting problems', exacerbates the issue. 

    \noindent\textit{\textbf{Entity misalignment}}: This issue arises when the retrieval mechanism accurately finds data but incorrectly links it to an entity in the knowledge graph, causing responses to deviate from the user's context. For example, in UNIX, a term like `read' might refer to a command or a configuration file, leading to misassociations if not accurately disambiguated. 
    
    \noindent\textit{\textbf{Composite query conundrums}}: This problem occurs when a user's query involves multiple issues, and the retrieval system typically focuses primarily on one, neglecting the others. For example, on platforms like AskUbuntu, a user might ask about `memory and CPU usage', and the system might only address the memory part.
    
    \noindent\textit{\textbf{Factual fidelity fallacies}}: This issue arises when the RAG system, skilled in retrieval and generation, delivers answers that lack factual accuracy or are outdated, a common issue in rapidly evolving platforms like AskUbuntu. For instance, a query about a software tool may elicit a response based on outdated versions. 
    
    \noindent\textit{\textbf{Contextual content crux}}: When the system retrieves broad or limited contents, it can produce answers that lack depth or specificity, a challenge often seen on platforms like AskUbuntu. For example, a query about a specific Ubuntu feature might get a general response if the corpus lacks in-depth content. 

\if{0}\begin{table}[]
\small
\scalebox{0.84}{
\begin{tabular}{|llll|}
\hline
\multicolumn{1}{|c|}{\textbf{Method}}                                                                       & \multicolumn{1}{l|}{\textbf{AskUbuntu}} & \multicolumn{1}{l|}{\textbf{Unix}} & \textbf{ServerFault} \\ \hline
\multicolumn{4}{|c|}{\textit{\textbf{Based on embedding algorithms}}}                                                                                                                                      \\ \hline
\multicolumn{1}{|l|}{\textbf{\begin{tabular}[c]{@{}l@{}}[w/o INST]~\textsc{TextGen} (Sentence BERT)\end{tabular}}}         & \multicolumn{1}{c|}{0.808}                   & \multicolumn{1}{c|}{0.783}              &      \multicolumn{1}{c|}{0.803}                \\ \hline
\multicolumn{1}{|l|}{\textbf{\begin{tabular}[c]{@{}l@{}}[w/o INST]~\textsc{TextGen} (BGE)\end{tabular}}}       & \multicolumn{1}{c|}{0.812}                   & \multicolumn{1}{c|}{0.809}              &    \multicolumn{1}{c|}{0.810}                  \\ \hline
\multicolumn{4}{|c|}{\textit{\textbf{Based on one-shot}}}                                                                                                                                      \\ \hline
\multicolumn{1}{|l|}{\textbf{\begin{tabular}[c]{@{}l@{}}[w/o INST]~\textsc{TextGen}\end{tabular}}}         & \multicolumn{1}{c|}{0.809}                   & \multicolumn{1}{c|}{0.807}              &                \multicolumn{1}{c|}{0.807}      \\ \hline
\multicolumn{1}{|l|}{\textbf{\begin{tabular}[c]{@{}l@{}}[w/o INST]~\textsc{GraphGen}\end{tabular}}}       & \multicolumn{1}{c|}{0.818}                   & \multicolumn{1}{c|}{0.809}              &    \multicolumn{1}{c|}{0.812}                  \\ \hline
\multicolumn{4}{|c|}{\textit{\textbf{Based on only knowledge graph context}}}                                                                                                                                      \\ \hline
\multicolumn{1}{|l|}{\textbf{\begin{tabular}[c]{@{}l@{}}[w/o INST]~\textsc{OnlyContextGen}[FT]\end{tabular}}}         & \multicolumn{1}{c|}{0.819}                   & \multicolumn{1}{c|}{0.807}              &   \multicolumn{1}{c|}{0.819}                   \\ \hline
\multicolumn{1}{|l|}{\textbf{\begin{tabular}[c]{@{}l@{}}[w/o INST]~\textsc{OnlyContextGen}[FG]\end{tabular}}}       & \multicolumn{1}{c|}{0.827}                   & \multicolumn{1}{c|}{0.820}              &    \multicolumn{1}{c|}{0.818}                  \\ \hline
\end{tabular}
}
\caption{Ablation study highlighting the performance of various methods. FT: From text, FG: From graph}
\label{tab:ablation}
\end{table}
\fi

\section{Complexity analysis}
As the dimension is fixed, the cosine similarity between two embeddings is $\bigO(1)$. For training set questions (say $n_q$ = variable), it becomes $n_q$ $\times$ $\bigO(1)$ = $\bigO(n_q)$. By including the query in the graph, pagerank becomes $\bigO((n_q + 1)+(e_q + e’))$ = $\bigO(n_q + 1 + e_q + e’)$ = $\bigO(n_q + e_q + e’)$. Assuming $e’$ can be at $max(n_q)$, it simplifies to $\bigO(n_q + n_q + e_q)$ = $\bigO(n_q + e_q)$. Therefore, the total complexity until pageRank calculation for each test instance is $\bigO(n_q + e_q)$. The worst-case complexity could be $\bigO(n_q + n_q^2)$ = $\bigO(n_q^2)$. For each dataset, $n_q$ and $e_q$ remain constant for all test instances.

\section{Conclusion}

This study addresses a notable challenge in CQA platforms: the automatic generation of answers across popular platforms often lacks clear problem definitions, leading to issues with proper knowledge grounding and factually incoherent responses. We present \textsc{GraphContextGen}, which, to the best of our knowledge, is the first solution that uses graph retrieval combined with knowledge graph context for this challenge in domain-specific CQA platforms. Our evaluations indicate that this model outperforms previous prominent approaches. Notably, human evaluators determine that answers generated by \textsc{GraphContextGen} exhibit greater factual coherence and knowledge grounding. We further demonstrate that researchers with constrained GPU resources can adopt this solution with smaller parameter LLMs and achieve performance that are at par with larger models.



\section{Limitation}

Using knowledge graphs with LLMs to generate technical answers encounters several challenges. The effectiveness of this approach partially depends on the accuracy and completeness of the knowledge graphs, as any gaps or inaccuracies can mislead the LLM's responses. LLMs tailored for summarization may sacrifice detail for conciseness, potentially overlooking critical nuances of technical topics. Challenges in handling ambiguous contexts and the potential for biases introduced during knowledge graph construction and LLM training further complicate accurate answer generation. Moreover, the computational resources required for processing large-scale knowledge graphs pose scalability issues. Ensuring that knowledge graphs remain up-to-date and that LLMs can adapt to new information without frequent retraining are ongoing concerns. Finally, while LLMs are versatile, they may not achieve the level of specificity and accuracy provided by systems specialized in particular non niche domains.

\section{Ethical consideration}

The information in our dataset is free from harmful or offensive materials. We take serious measures to anonymize and handle any personal or sensitive data with the highest level of confidentiality. Protection of participants' privacy is our priority and we consistently ensure to acquire their informed consent when collecting, annotating, and analyzing data. We provide equal incentives to all annotators for their efforts toward the annotation task.

\bibliography{custom}
\bibliographystyle{acl_natbib}

\appendix

\section{Appendix}

\subsection{Instruction tune hyperparameters}
\label{sec:insttunehyp}
\begin{table}[h]
\centering
\small
\begin{tabularx}{0.45\textwidth}{|l|X|}
\hline
\textbf{Hyperparameter} & \textbf{Value} \\
\hline
Learning Rate & 2e-4\\
\hline
Batch Size & 4\\
\hline
Gradient Accumulation Step & 1\\
\hline
Number of Epochs & 10\\
\hline
Weight Decay & 0.001\\
\hline
Optimizer & {paged adamw 32bits}\\
\hline
LR scheduler & {cosine}\\
\hline
Warmup ratio & 0.03\\
\hline
Max grad norm & 0.3\\
\hline
bf16 & True\\
\hline
LoRA r, alpha, dropout & 64, 16, 0.1\\
\hline
bf16 & True\\
\hline
Quantization & 4bit\\
\hline
PEFT Techniques & LoRA\\
\hline
Trainer & SFTT\\
\hline
\end{tabularx}
\caption{Hyperparameters for instruction tuning the LLM using SFTT trainer.}
\end{table}

\begin{table*}[htb]
\centering
\tiny
\begin{tabular}{|l|l|l|}
\hline
\multicolumn{1}{|c|}{\textbf{Question}}                                                                                                                                                                                                                                                                                                                                                                                                                                                 & \multicolumn{1}{c|}{\textbf{Related Question from PPR}}                                                                                                                                                                                                                                                                                                                                                                                                                                                    & \multicolumn{1}{c|}{\textbf{Related Question from simple similarity}}                                                                                                                                                                                                                                                                                                                                                                                                                                 \\ \hline
\begin{tabular}[c]{@{}l@{}}Experiencing Intermittent Network Failures \\ on Ubuntu Server After Recent Update. \\ \\ \textbf{Desc:} After a recent update on my Ubuntu \\ 20.04 server,  I'm experiencing intermittent \\ network failures. The server loses connectivity\\ randomly, and I've been unable to diagnose the \\ issue. Here's the output of \texttt{ifconfig} and \\ \texttt{dmesg | grep eth0} after the failure occurs...\end{tabular}                                                    & \begin{tabular}[c]{@{}l@{}}How can I rollback a recent Ubuntu update to \\ troubleshoot network connectivity issues? \\ \\ \textbf{Desc:} Following recent network issues on \\ my Ubuntu server, I suspect a recent update \\ might be the cause. I need to rollback this \\ update to confirm. What is the safest way to \\ revert the last system update? Is there a way \\ to identify which packages were updated \\ and selectively rollback, or do I need to restore \\ from a backup?\end{tabular} & \begin{tabular}[c]{@{}l@{}}What are some common network troubleshooting \\ tools in Ubuntu for diagnosing connectivity problems? \\ \\ \textbf{Desc:} In dealing with intermittent network failures on \\ my Ubuntu server, I'm looking for effective tools or \\ commands to diagnose the issue. What are the best tools\\ available in Ubuntu for network troubleshooting, especially\\ for monitoring and logging network activity over time to \\ catch these intermittent failures?\end{tabular} \\ \hline
\begin{tabular}[c]{@{}l@{}}Script for Automating Log File Rotation \\ and Compression Not Working as Expected\\ on Linux. \\ \\ \textbf{Desc:} I'm attempting to create a bash script\\ to automate log file rotation and compression \\ in a Linux environment. The script is supposed\\ to find all log files under \texttt{/var/log} , compress them, \\ and then move them to \texttt{/archive/logs}.However, \\ it's not working as expected, and some log files \\ are being missed...\end{tabular} & \begin{tabular}[c]{@{}l@{}}How can I set up a cron job to run this script\\ daily at midnight? \\ \\ \textbf{Desc:} I have a script for log file rotation and\\ compression, but I'm not sure how to set it up\\ as a cron job to run automatically. What is the \\ correct way to schedule this script in cron to \\ run daily at midnight? Are there any specific \\ considerations for running such scripts as cron jobs?\end{tabular}                                                                  & \begin{tabular}[c]{@{}l@{}}What are the best practices for managing log files in\\ a Unix environment?\\  \\ \textbf{Desc:} As I work on automating log file rotation and\\  compression, I want to ensure I'm following best \\ practices. What are the recommended strategies\\  for log file management in a Unix environment? \\ This includes considerations for log rotation frequency,\\  compression, archiving, and ensuring log integrity and security?\end{tabular}                        \\ \hline
\end{tabular}
\caption{Retrieved questions from PPR (graph structure based) and simple similarity for certain questions.}
\label{tab:pprsample}
\end{table*}

\begin{table*}[htb]
\centering
\small
\scalebox{0.70}{
\begin{tabular}{|l|l|l|} 
\hline
\textbf{ID} & \textbf{Error Type}           & \textbf{Samples}                                                                                                                                                                                                                                                                                                                                                                                                                                                                                                                                                                          \\ 
\hline
\textbf{1}  & Misaligned Retrieval Outcomes & \begin{tabular}[c]{@{}l@{}}Question: How can I change the \colorbox{green!15}{desktop environment} in Ubuntu?\\ Retrieved Content: Steps to change the \colorbox{red!15}{desktop wallpaper} in Ubuntu.\\ Generated Answer: To \colorbox{red!15}{change the wallpaper}, right-click on the desktop and select 'Change Wallpaper'........\\ Analysis: The retrieved content precisely discusses \colorbox{red!15}{changing the wallpaper}, but the user's query was about \\ \colorbox{green!15}{changing the entire desktop environment}, not just the wallpaper.\end{tabular}                                                                                                                \\ 
\hline
\textbf{2}  & Entity Misalignment           & \begin{tabular}[c]{@{}l@{}}Question: What is the use of  \colorbox{green!15}{\`{}chmod\`{}} in \colorbox{green!15}{UNIX}?\\ Retrieved Entity: \colorbox{red!15}{\`{}chown\`{}} command details from the \colorbox{red!15}{KG}.\\ Generated Answer: ..........\colorbox{red!15}{\`{}chown\`{}} is used to change the owner of a file or directory...................................\\ Analysis: The retrieved information is accurate about \colorbox{red!15}{\`{}chown\`{}}, but the user's query was about \colorbox{green!15}{\`{}chmod\`{}}. \\ This represents a wrong entity linking.\end{tabular}                                                                                                                              \\ 
\hline
\textbf{3}  & Composite Query Conundrums    & \begin{tabular}[c]{@{}l@{}}Question: How do I \colorbox{green!15}{configure the firewall} on my server to \colorbox{green!15}{allow SSH access}, and also \colorbox{green!15}{set up an FTP server}\\ for file transfers?\\ Retrieved Content: Detailed steps on \colorbox{green!15}{configuring the firewall} for \colorbox{green!15}{SSH access}.\\ Generated Answer: To \colorbox{green!15}{configure the firewall} for \colorbox{green!15}{SSH access}, you need to open port 22 and apply the necessary rules...\\ Analysis: While the retrieved content and generated answer correctly address the \colorbox{green!15}{firewall configuration} \\ for \colorbox{green!15}{SSH}, the question's second aspect about \colorbox{red!15}{setting up an FTP server} is entirely neglected.\end{tabular}  \\ 
\hline
\textbf{4}  & Factual Fidelity Fallacies    & \begin{tabular}[c]{@{}l@{}}Question: How can I \colorbox{green!15}{upgrade} to the \colorbox{green!15}{latest version} of \colorbox{green!15}{Ubuntu}?\\ Retrieved Content: Steps to \colorbox{green!15}{upgrade} \colorbox{green!15}{Ubuntu} 16.04 to \colorbox{green!15}{Ubuntu} \colorbox{red!15}{18.04}.\\ Generated Answer: To \colorbox{green!15}{upgrade} to the \colorbox{green!15}{latest version} of \colorbox{green!15}{Ubuntu}, follow these steps to upgrade from Ubuntu 16.04 to \colorbox{red!15}{18.04}...\\ Analysis: While the retrieved and generated content might be factually correct for the specific version transition, it's \\ outdated if the latest version is \colorbox{red!15}{beyond Ubuntu 18.04}.\end{tabular}                                                                                                 \\ 
\hline
\textbf{5}  & Contextual Content Crux       & \begin{tabular}[c]{@{}l@{}}Question: How do I solve the \colorbox{green!15}{XYZ error} in Ubuntu?\\ Retrieved Content: A post briefly mentioning the \colorbox{green!15}{XYZ error} with a comment saying, \colorbox{red!15}{"Ï also faced this."}\\ Generated Answer: The \colorbox{green!15}{XYZ error} is an issue \colorbox{red!15}{some users have faced} in Ubuntu..........................................\\ Analysis: The retrieved content acknowledges the error but \colorbox{red!15}{provides no solution} or detailed information,\\ leading to an unsatisfactory and unhelpful answer.\end{tabular}                                                                                                 \\
\hline
\end{tabular}
}
\caption{The actual rationale being marked with the \colorbox{green!15}{green} and retrieved and generated rationale marked as \colorbox{red!15}{red}.}
\label{tab:erroranalysis}
\end{table*}

\begin{table*}[htb]
\centering
\scalebox{0.62}{
\begin{tabular}{|l|l|}
\hline
Question & How do I extract an ISO file?                                                                                                                                                                                                                                                                                                                                                                                                                                                                                                                                                                                                                                                                                                                                                                                                                                                                                                                                                                                                                                                                                                                                                                                                                                                                                                                                                                                                                                                                                                                                                                                                                                                                                                                                                                                                                                                                                                                                                                                                                                                                                                                                                                                                                                                                                                                                                          \\ \hline
Context  & \begin{tabular}[c]{@{}l@{}}Question: Has anyone used isoInfo to extract ISO files? How does it compare to others?\\ \\ Answer: Hey! So, `isoInfo` isn't for extracting ISO files; it just shows details about them. You wanna extract? \\ Use `isoinfo` from the same package. Like this: isoinfo -i image.iso -x /PATH/INSIDE/ISO \textgreater output.file\\ Now, comparing `isoinfo` with other tools:\\ 1. 7-Zip: Cool for many file types, has both GUI and command-line. \\ 2. WinRAR: Good for ISOs, but it's paid (though many stick to the trial).\\ 3. Daemon Tools: Great for 'mounting' ISOs, but can extract too.\\ 4. dd \& mount (Linux stuff): More techie, but pretty powerful.\\ If you're a Linux person comfy with commands, go for `isoinfo`. If not, maybe try a tool with a friendly interface. \\ \\ Question: Can you recommend a good free tool for extracting ISO files on Ubuntu 20.04?\\ \\ \\ Answer: Hey! On Ubuntu 20.04 and wanna extract ISO files? Easy peasy:\\ 1. Archive Manager: Built-in tool. Just right-click the ISO, hit "Open With Archive Manager", and boom, you're in.\\ 2. 7-Zip: Not just a Windows thing! Get the Linux version, `p7zip`, with:\\  sudo apt install p7zip-full\\ Then run: 7z x file.iso\\ 3. mount: A bit techie. Mount the ISO, copy its stuff, like:\\ mkdir /path/to/mountpoint\\ sudo mount -o loop file.iso /path/to/mountpoint\\ cp -r /path/to/mountpoint /path/to/destination\\ sudo umount /path/to/mountpoint\\ 4. isoinfo: Fancy extracting specific stuff? Grab it with:\\ sudo apt install genisoimage\\ Most folks will find Archive Manager or 7-Zip super easy. But if you're a command line junkie, try `mount` or `isoinfo`.\\ \\ Question: What could be the important context to answer this?\\ Answer:\\ ('Archive Manager is a built-in tool'),\\  ('Archive Manager is used for extracting ISO files'),\\  ('7-Zip is a Linux version'),\\  ('7-Zip is used to extract ISO files'),\\  ('7-Zip is used for extracting ISO files'),\\  ('WinRAR is a paid tool'),\\  ('Daemon Tools is used for mounting ISOs'),\\  ('dd \& mount is a Linux stuff'),\\  ('isoinfo is used for extracting specific stuff'),\\  ('isoinfo is used for extracting ISO files'),\\  ('mount is used for extracting ISO files'){}\end{tabular} \\ \hline
Answer   & \begin{tabular}[c]{@{}l@{}}To extract an ISO in Linux/Ubuntu:\\ 1. Archive Manager: Built-in. Right-click the ISO, select "Open With Archive Manager", and extract.  \\ 2. 7-Zip: Install using `sudo apt install p7zip-full` and extract with `7z x file.iso`. \\ 3. mount: Create a mount point `mkdir /mount/spot`. Mount with `sudo mount -o loop file.iso /mount/spot`, copy with \\ `cp -r /mount/spot /destination`, then unmount `sudo umount /mount/spot`.  \\ 4. isoinfo: Install with `sudo apt install genisoimage`. \\ To extract specifics, use `isoinfo -i image.iso -x /PATH/INSIDE/ISO \textgreater output.file`.\\ For ease, go with Archive Manager or 7-Zip. If you're command-line savvy, try `mount` or `isoinfo`.\end{tabular}                                                                                                                                                                                                                                                                                                                                                                                                                                                                                                                                                                                                                                                                                                                                                                                                                                                                                                                                                                                                                                                                                                                                                                                                                                                                                                                                                                                                                                                                                                                                                                                                                                  \\ \hline
\end{tabular}
}
\caption{Sample prompt an generated answer.}
\label{tab:sampleexample}
\end{table*}

\subsection{Pre-LLM era baselines}
\label{sec:prellmbase}
\subsubsection{\textbf{AnswerBot~\cite{Xu:2017, Cai:2019}:}}
Authors of this work proposed an approach called AnswerBot, where the task is to generate a summary from diverse answers for a query. They followed three major steps -- relevant question retrieval, useful answer paragraph selection, diverse answer summary generation. For retrieval, they used word2vec model and relevance calculation algorithm. For answer paragraph selection, they used various query, paragraph and user related features --~\emph{relevance to query}, ~\emph{entity overlap}, ~\emph{information entropy}, ~\emph{semantic pattern}, ~\emph{format patterns}, ~\emph{paragraph position}, ~\emph{vote on answer}. In answer summary generation stage, they used maximal marginal relevance (MMR) algorithm to select a subset of answer paragraphs. Further they used selected answer paragraphs to form the answer summary. 
\subsubsection{\textbf{GenQA~\cite{hsu:2021}:}} Authors of this paper proposed a framework to generate answers from the top candidates of a set of answer selection models. Instead of selecting the best candidates, they train a sequence to sequence transformer model to generate an answer from candidate set.
\subsubsection{~\textbf{TechSumBot~\cite{10172591}:}} Authors of this paper show that developers frequently turn to StackOverflow for solutions, but they often encounter redundant or incomplete results. Current tools designed to summarize StackOverflow answers have clear drawbacks: they predominantly depend on manually-designed features, they struggle to filter out repetitive content, and they usually target specific programming languages. This tool autonomously produces answer summaries by extracting and ranking answers for their relevance, measuring the core importance of each sentence, and eliminating redundant details. Presented in a search engine format, TechSumBot's efficiency is benchmarked against existing StackOverflow summary methods.


\subsection{Sample prompt and generated answer}
The sample prompt and the generated answer for a specific example is shown in Table~\ref{tab:sampleexample}.

\subsection{Sample questions retrieved from our method}
We include a few examples in Table~\ref{tab:pprsample} that take into account both PPR (graph structure-based) and simple similarity for certain questions. The questions retrieved by the PPR method are very specific, to-the-point, and strongly related to the actual question. The simple similarity-based questions, on the other hand, are very generic (e.g., What are some common network troubleshooting tools…, What are the best practices for managing log files…).

\subsection{Enhanced context formulation}
  \begin{figure}
  \centering
    \includegraphics[scale=0.50]{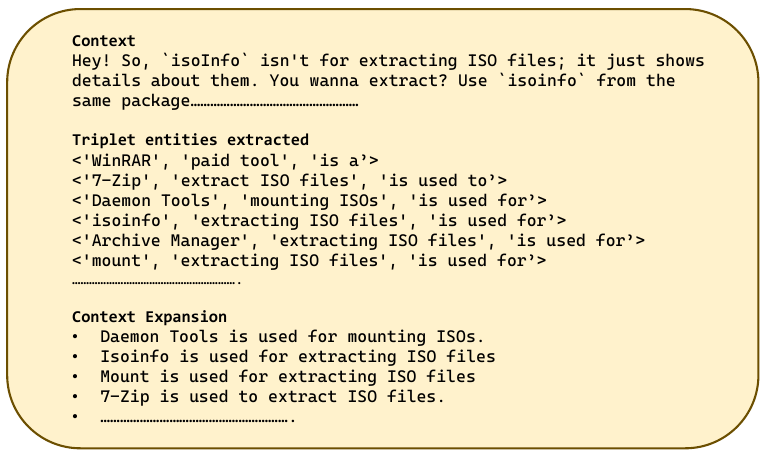} 
    \caption{Sample context preparation.}\label{fig:enhCon}
  \end{figure}

\subsection{Human annotation}
\label{appd:humanannotation}
We engaged nine undergraduate students, each an expert in their domain, to undertake our annotation task, dividing them into groups of three for each dataset. All these students are majoring in Computer Science and have a proven track record of contributing high-quality answers within relevant community platforms. They voluntarily joined our project after receiving an invitation through departmental emails and were rewarded with Amazon gift cards for their contributions. Each undergraduate student annotates 50 instances individually. We utilize the Doccano\footnote{https://elements.heroku.com/buttons/doccano/doccano} tool for obtaining the annotations. The annotators provided feedback on a scale from 1 to 5. A rating of `1' means the answer is unhelpful or misleading, while a `5' indicates an exemplary response. In Figure~\ref{fig:humanfeedback}, we display the feedback distribution for test instances. In the pie chart, the outer ring represents the three datasets. For every dataset, five segments in the inner ring depict the distribution of ratings from 1 to 5. The plot reveals that the answers generated by our model for both the AskUbuntu and the Unix test cases predominantly have a rating of 4 as per human judgement, while those generated by our model for the ServerFault test cases predominantly have ratings of 5. Next we compute `win rate' which refers to the percentage of individuals who favor the output from our model over the standard zero-shot output. In our analysis comparing answers generated by our model with those from a simple zero-shot approach, we observe a notable trend in win rates across the three platforms. Specifically, for Askubuntu, Unix, and ServerFault, the win rates are 58\%, 63\%, and 53\%, respectively. These rates consistently exceed the 50\% benchmark.







\end{document}